\documentclass[conference]{IEEEtran}
\IEEEoverridecommandlockouts
\usepackage{cite}
\usepackage{amsmath,amssymb,amsfonts}
\usepackage{algorithmic}
\usepackage{graphicx}
\usepackage{textcomp}
\usepackage{xcolor}
\usepackage{algorithm}
\usepackage{bm}
\usepackage{subfigure}
\usepackage{multirow}
\usepackage{booktabs}
\usepackage{array}
\usepackage{makecell}
\usepackage{url} % 确保导入了 url 包
\usepackage{booktabs}
\usepackage[justification=centering]{caption}

\usepackage[defaultcolor=magenta]{changes}
\usepackage{subcaption}
\usepackage[hidelinks,bookmarks=false]{hyperref}

\def\BibTeX{{\rm B\kern-.05em{\sc i\kern-.025em b}\kern-.08em
    T\kern-.1667em\lower.7ex\hbox{E}\kern-.125emX}}

\begin{document}

\title{PWC-MoE: Privacy-Aware Wireless Collaborative Mixture of Experts}

\author{\IEEEauthorblockN{Yang~Su\IEEEauthorrefmark{1},\ Na~Yan\IEEEauthorrefmark{1},\ Yansha~Deng\IEEEauthorrefmark{1}, and Robert Schober \IEEEauthorrefmark{2}}
	\IEEEauthorblockA{
		\IEEEauthorrefmark{1}Department of Engineering, 
		King's College London, London, UK
        }
        \IEEEauthorblockA{
            \IEEEauthorrefmark{2}Institute for Digital Communication, 
		Friedrich-Alexander-Universit\"at Erlangen-N\"urnberg, Erlangen, Germany
	}	
}

\maketitle
\begin{abstract}
Large language models (LLMs) hosted on cloud servers alleviate the computational and storage burdens on local devices but raise privacy concerns due to sensitive data transmission and require substantial communication bandwidth, which is challenging in constrained environments. 
In contrast, small language models (SLMs) running locally enhance privacy but suffer from limited performance on complex tasks. To balance computational cost, performance, and privacy protection under bandwidth constraints, we propose a privacy-aware wireless collaborative mixture of experts (PWC-MoE) framework.
Specifically, PWC-MoE employs a sparse privacy-aware gating network to dynamically route sensitive tokens to privacy experts located on local clients, while non-sensitive tokens are routed to non-privacy experts located at the remote base station. To achieve computational efficiency, the gating network ensures that each token is dynamically routed to and processed by only one expert. To enhance scalability and prevent overloading of specific experts, we introduce a group-wise load-balancing mechanism for the gating network that evenly distributes sensitive tokens among privacy experts and non-sensitive tokens among non-privacy experts. To adapt to bandwidth constraints while preserving model performance, we propose a bandwidth-adaptive and importance-aware token offloading scheme. This scheme incorporates an importance predictor to evaluate the importance scores of non-sensitive tokens, prioritizing the most important tokens for transmission to the base station based on their predicted importance and the available bandwidth.
Experiments demonstrate that the PWC-MoE framework effectively preserves privacy and maintains high performance even in bandwidth-constrained environments, offering a practical solution for deploying LLMs in privacy-sensitive and bandwidth-limited scenarios.
\end{abstract}


\section{Introduction}
Large language models (LLMs) have exhibited exceptional performance in understanding and generating natural language across a wide range of tasks, such as applications in chatbots and search engines\cite{zhao2023survey}. However, state-of-the-art LLMs, such as ChatGPT, typically have billions of parameters, requiring substantial computational and storage resources for deployment, which makes it challenging to run them locally on client devices. As a result, LLMs are primarily deployed at cloud servers, with client devices accessing them remotely. While this approach addresses the resource limitations of local devices, it introduces significant privacy concerns, as user data including sensitive information will need to be transmitted to and processed on external servers. 

To address the challenges of resource demand and privacy concerns, small language models (SLMs) \cite{van2024survey} have been proposed as an alternative to LLMs. These smaller models are often derived from LLMs through techniques such as model distillation. For example, the authors in \cite{jiao-etal-2020-tinybert} proposed TinyBERT, a SLM distilled from BERT, which effectively transfers knowledge from a large ``teacher" model to a smaller ``student" model. By leveraging such distillation techniques, SLMs can significantly reduce computational and storage requirements while maintaining competitive performance. However, despite these advancements, there remains a noticeable performance gap between SLMs and their larger counterparts, particularly for complex tasks.

To address the performance gap between SLMs and LLMs, authors in \cite{zhang2024llm} introduced an LLM cascade strategy with multi-objective optimization, where user input queries are first processed by a local SLM and routed to a server-side LLM only when necessary. This approach takes into account multiple factors, such as computational cost, performance, and privacy, when determining whether to process a query locally or offload it to the server.
However, while this method improves privacy by allowing some queries to remain on the local device, sensitive information may still be sent to the server when the local model cannot handle the query. Additionally, the binary nature of the decision-making process (local or server-side) limits the flexibility 
and granularity of privacy control.

Existing cascading systems often overlook bandwidth efficiency in resource-constrained networks. For example, while the LLM cascade strategy in \cite{zhang2024llm} reduces computational costs by selectively routing queries, it does not explicitly address the communication overhead associated with transmitting data between client devices and remote servers. In bandwidth-constrained scenarios, the transmission of large amounts of data can lead to significant delays, increased costs, and degraded user experience.

To address these issues, in this paper, we propose a privacy-aware wireless collaborative mixture of experts (PWC-MoE) framework, which integrates privacy preservation and bandwidth efficiency into the deployment of LLMs in wireless collaborative systems. The framework is built upon the MoE architecture\cite{cai2025survey}, a scalable and efficient approach for implementing LLMs that dynamically activates only a subset of model parameters for each input, significantly reducing computational costs while maintaining high performance.
The main contributions of this paper are summarized as follows:
\begin{itemize}
    \item We propose a PWC-MoE framework that incorporates a sparse privacy-aware gating network to dynamically route tokens to the most appropriate experts. Sensitive tokens are processed locally by privacy experts on client devices, while non-sensitive tokens are handled by non-privacy experts at the remote base station. To ensure balanced utilization of privacy and non-privacy experts, we introduce a group-wise load balancing strategy that enforces uniform token distribution within each expert group, improving efficiency and scalability.

    \item We propose a bandwidth-adaptive and importance-aware token offloading mechanism that leverages an importance predictor to evaluate and rank non-sensitive tokens based on their contribution to the model's output. During deployment, the ranked tokens are dynamically selected for transmission according to the available bandwidth, ensuring that the most important tokens are sent to the base station with higher priority.

    \item We conduct experiments using a privacy-aware MoE model built upon GPT-2, fine-tuned on the Banking77 classification dataset. Our results demonstrate that the model achieves stable convergence during training, while the importance predictor-based token selection method outperforms baseline strategies, showing comparable accuracy with significantly fewer transmitted tokens.
\end{itemize}

The rest of the paper is organized as follows: Section II introduces the system model; Section III details our novel privacy-aware MoE model; Section IV discusses our bandwidth-adaptive and importance-aware token offloading scheme; Section V presents the numerical results; and finally, Section VI concludes the paper.

\section{System Model}
\begin{figure}[htbp]
\centerline{\includegraphics[scale=0.47]{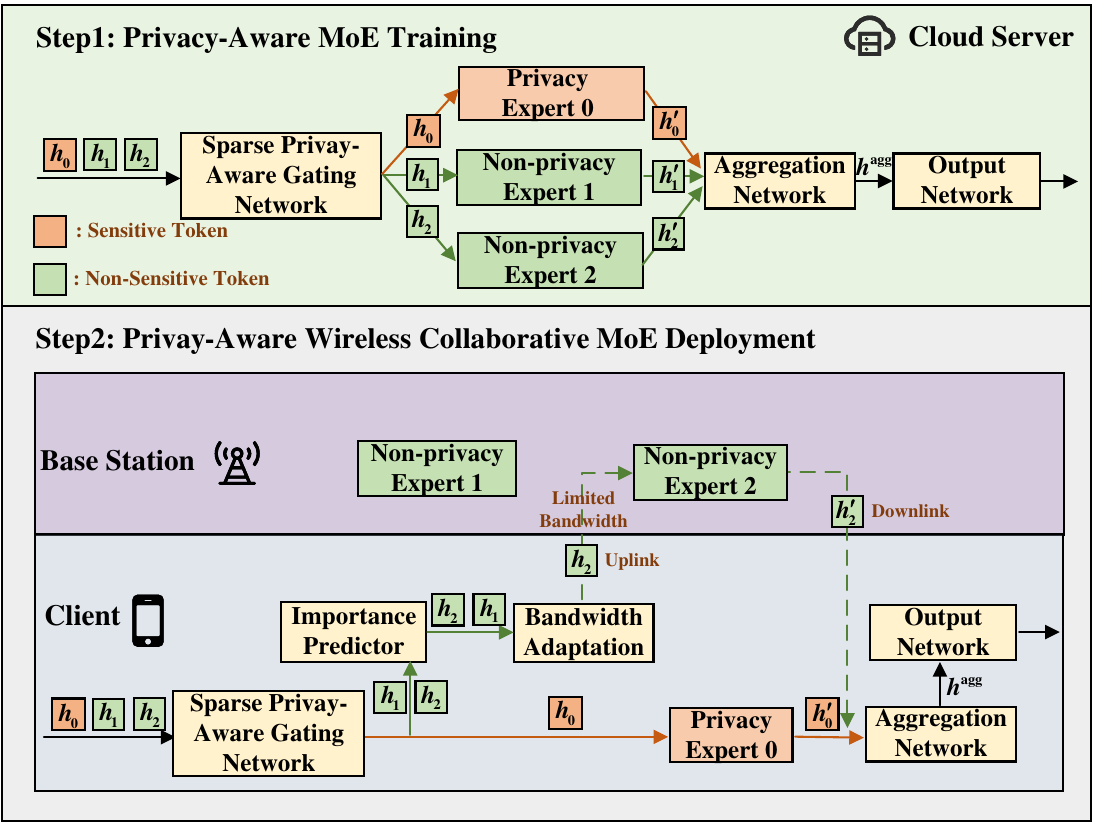}}
\caption{PWC-MoE framework.}
\label{system_model}
\end{figure}

As shown in Fig.~\ref{system_model}, we propose a PWC-MoE framework that builds on a novel privacy-aware MoE model and introduces a collaborative deployment strategy. The framework consists of a remote base station and a local client.
When a query is input to the client, a privacy-aware gating network dynamically routes tokens based on their sensitivity. Sensitive tokens \(\mathcal{T}_\text{s}\) are processed locally by privacy experts, ensuring private information remains on the client. Non-sensitive tokens \(\mathcal{T}_{\text{ns}}\) are filtered by selecting a subset of the most important tokens, \(\mathcal{I} \subseteq \mathcal{T}_\text{ns}\), subject to bandwidth constraints. This subset \(\mathcal{I}\) is transmitted to the base station, for processing by non-privacy experts.
The base station then returns the processed results to the client, which aggregates outputs from both local privacy experts and remote non-privacy experts to generate the final output.

\subsection{Privacy-Aware MoE Training}
The proposed privacy-aware MoE model is trained at a cloud server, as detailed in Section III. The model features a sparse privacy-aware gating network that employs a Top-1 selection mechanism, dynamically routing each token to the most appropriate expert. This mechanism ensures that sensitive tokens \(\mathcal{T}_\text{s}\) are directed to privacy experts, while non-sensitive tokens \(\mathcal{T}_\text{ns}\) are handled by non-privacy experts. To address potential imbalances in token distribution and prevent overloading specific experts, the gating network incorporates load balancing strategies that distribute sensitive tokens among privacy experts and non-sensitive tokens among non-privacy experts, effectively managing workloads within each expert group. After processing by the selected experts, the outputs are passed to an aggregation network, which computes a weighted sum of the outputs from all experts. The aggregated result is then passed through a task-specific output network, which transforms the representation into the desired output format, such as class probabilities for classification tasks, or sequences for generation tasks.

\subsection{Collaborative Deployment in the PWC-MoE Framework}
To achieve an optimal balance between computational cost, performance, and privacy protection, a collaborative deployment strategy is proposed. In this strategy,
privacy experts from the privacy-aware MoE model are deployed locally on clients to process sensitive tokens, while non-privacy experts are hosted at the remote base station to handle non-sensitive tokens. To address bandwidth constraints, an importance predictor is employed to evaluate the significance of non-sensitive tokens (described in Section IV), and only a subset of the most important tokens \(\mathcal{I}\), is transmitted to the base station for processing.

\subsubsection{Uplink Communication Model}
We consider a common urban scenario where the client  has no line-of-sight (NLoS) to the base station\cite{3gpp_tr_38_901_v16_1_0}, the path loss is modeled as  
\begin{equation}
    {PL} = 32.4 + 20\log_{10}(f_\text{c}) + 30\log_{10}(d_\text{c}),
    \label{eq:pl}
\end{equation}
where \( f_\text{c} \)  is the carrier frequency, and \( d_\text{c} \) is the distance between the client and the base station.

Incorporating shadowing and small-scale fading effects, the overall channel gain \( h^{\text{ul}} \) is given by  
\begin{equation}
    h^{\text{ul}} = 10^{-{PL}/10} \, \psi \, \chi,
    \label{eq:channel_gain}
\end{equation}
where the large-scale shadowing effect is modeled as  
\begin{equation}
\psi = 10^{\xi/10}, \quad \xi \sim \mathcal{N}(0, \sigma^2),
\end{equation}
and the small-scale rayleigh fading component is  
\begin{equation}
\chi = |\eta|^2, \quad \text{with } \eta \sim \mathcal{CN}(0,1).
\end{equation}

The Signal-to-Noise Ratio (SNR) for the client is given by:
\begin{equation}
\text{SNR}^{\text{ul}} = \frac{P h^{\text{ul}}}{N_0 W_\text{c}},
\end{equation}
where $P$ is the transmit power of the client, $N_0$ is the noise power spectral density (PSD), and $W_\text{c}$ is the limited bandwidth allocated to client.

The uplink transmission rate \( R^{\text{ul}} \) for the client can be expressed as:
\begin{equation}
R^{\text{ul}} = W_c \cdot \log_2 \left( 1 + \text{SNR}^{\text{ul}} \right).
\end{equation}

Each token requires \( b_{\text{token}} \) bits for representation. Given the uplink communication time \( T^{\text{ul}} \), the maximum number of tokens \( m^{\text{ul}} \) that can be transmitted is given by:
\begin{equation}
m^{\text{ul}} = \frac{T^{\text{ul}} \cdot R^{\text{ul}}}{b_{\text{token}}}.
\end{equation}

\subsubsection{Downlink Communication Model}
In the proposed framework, the processed results returned by the base station maintain the same dimensions as the input token embeddings, as the expert processing does not modify the size of the token embeddings. Given the base station's significantly higher transmission power and sufficient resources, we assume that the downlink communication system can reliably transmit these results.

\subsection{Problem Formulation}
The objective of the proposed PWC-MoE framework is to maximize the overall system performance (e.g., classification accuracy) by selecting the most important non-sensitive tokens for uplink transmission to the base station, where they are processed by non-privacy experts. Meanwhile, the number of uploaded tokens must not exceed the bandwidth constraints. This can be formulated as the following optimization problem:
\begin{align}
\max_{\mathcal{I} \subseteq \mathcal{T}_\text{ns}} \, & f(\mathcal{T}_\text{s}, \mathcal{I}) \\
\text{s.t.} \quad & |\mathcal{I}| \leq m^{\text{ul}},
\end{align}
where
\(\mathcal{T}_\text{s}\) represents the set of sensitive tokens processed locally by privacy experts on the client.
\(\mathcal{T}_\text{ns}\) represents the set of non-sensitive tokens, from which a subset \(\mathcal{I}\) is selected for uplink transmission to the base station.
\(|\mathcal{I}|\) denotes the number of tokens in the subset \(\mathcal{I}\).
\(m^{\text{ul}}\) is the maximum number of tokens that can be transmitted under the uplink bandwidth constraint.
\(f(\mathcal{T}_\text{s}, \mathcal{I})\) represents the system performance metric, which evaluates the contribution of both sensitive tokens \(\mathcal{T}_\text{s}\) (processed locally) and the selected non-sensitive tokens \(\mathcal{I}\) (processed at the base station) to the final prediction accuracy or confidence of the model.

\section{Privacy-aware Mixture of Experts}
In this section, we introduce our novel privacy-aware MoE model, along with the identification of privacy tokens, the design of the sparse privacy-aware gating network, and the loss functions employed for optimization.

\subsection{Identification of Privacy Tokens}
Privacy tokens are defined as tokens that represent sensitive information such as names, addresses, phone numbers, dates of birth, or other personally identifiable information (PII). In this work, we will use existing methods to identify privacy tokens, including rule-based approaches (e.g., regular expressions) or deep learning-based named entity recognition (NER) models\cite{akbik2019pooled}.  The input sequence is represented as
\begin{equation}
\mathbf{x} = [x_1, x_2, \dots, x_L],
\end{equation}
where \(x_i\) represents the \(i\)th token in the sequence, and $L$ is the total length of the sequence. After passing through the embedding layer, the sequence is transformed into embeddings
\begin{equation}
\mathbf{h} = [h_1, h_2, \dots, h_L],
\end{equation}
where \(h_i \in \mathbb{R}^d\) is the embedding of the \(i\)th token, and \(d\) is the dimensionality of the embedding space.

A binary privacy mask \(\mathbf{m} = [m_1, m_2, \dots, m_L]\) is constructed to indicate whether tokens in the input sequence \(\mathbf{x}\) are sensitive. The mask is defined as
\begin{equation}
m_i = 
\begin{cases} 
1, & \text{if } x_i \text{ is sensitive}, \\
0, & \text{otherwise}.
\end{cases}
\end{equation}

This mask ensures that privacy-sensitive operations are applied exclusively to the relevant tokens.

\subsection{Sparse Privacy-Aware Gating Network}  
To ensure that privacy tokens and non-privacy tokens are routed to different sets of experts in a way that is both efficient and scalable, we introduce a sparse privacy-aware gating mechanism.

\subsubsection{Gating Logits Computation}
For each token embedding \(h_i \in \mathbb{R}^d\), the gating logits \(\mathbf{g}_i \in \mathbb{R}^K\) are computed as
\begin{equation}
\mathbf{g}_i = \mathbf{W}_g h_i + \mathbf{b}_g,
\end{equation}
where \(\mathbf{W}_g \in \mathbb{R}^{K \times d}\) is the gating weight matrix, \(\mathbf{b}_g \in \mathbb{R}^K\) is the gating bias vector, and \(K\) is the total number of experts.

\subsubsection{Privacy Isolation Mechanism}  
Let the first \(K_\text{p}\) experts be designated as privacy experts, and the remaining \(K_\text{np} = K - K_\text{p}\) experts as non-privacy experts. For each token \(x_i\), based on the binary privacy mask \(m_i \in \{0, 1\}\), the gating logits \(\mathbf{g}_i\) are adjusted as follows:  
\begin{equation}  
\mathbf{g}_i' =  
\begin{cases}  
[\mathbf{g}_i[:K_\text{p}], -\infty, \dots, -\infty], & \text{if } m_i = 1, \\  
[-\infty, \dots, -\infty, \mathbf{g}_i[K_\text{p}:]], & \text{if } m_i = 0.  
\end{cases}  
\end{equation}
Here, for sensitive tokens, only the first \(K_\text{p}\) components of \(\mathbf{g}_i\) are retained, corresponding to the privacy experts, while the remaining \(K_\text{np}\) components are set to \(-\infty\). Similarly, for non-sensitive tokens, only the last \(K_\text{np}\) components of \(\mathbf{g}_i\) are retained, corresponding to the non-privacy experts, while the first \(K_\text{p}\) components are set to \(-\infty\).

\subsubsection{Gumbel-Softmax Operation} 
After obtaining the modified gating logits \(\mathbf{g}_i'\), we apply the Gumbel-Softmax operation \cite{DBLP:conf/iclr/JangGP17} to compute the expert selection probabilities for token $x_i$.
Specifically, the \(j\)th component of the output \(\mathbf{z}_i \in \mathbb{R}^K\) is computed as:  
\begin{equation}  
z_{i,j} = \frac{\exp\left((g_{i,j}' + \gamma_j) / \tau\right)}{\sum_{k=1}^K \exp\left((g_{i,k}' + \gamma_{l}) / \tau\right)},
\end{equation}
where \(g_{i,j}'\) is the \(j\)th component of the modified gating logits \(\mathbf{g}_i'\), each \(\gamma_j \sim \text{Gumbel}(0, 1)\) is independently sampled noise, and \(\tau > 0\) is the temperature parameter that controls the smoothness of the output distribution.
\addtolength{\topmargin}{0.03in}
The output \(\mathbf{z}_i = [z_{i,1}, z_{i,2}, \dots, z_{i,K}]\) is a probability vector, where \(z_{i,j} \in [0, 1]\) and \(\sum_{j=1}^K z_{i,j} = 1\).  
To enforce discrete selection, we apply hard Gumbel-Softmax with the straight-through estimator: the output is a one-hot vector
\begin{equation}
o_i = \text{one\_hot}\left(\arg\max_j \left(g_{i,j}' + \gamma_j\right)\right),
\end{equation}
while gradients are computed from the continuous \( \mathbf{z}_i \) to maintain differentiability.

\subsection{Weighted Aggregation Network}
After each token is processed by its assigned expert, the outputs from all tokens are collected into a sequence
\begin{equation}
\mathbf{h'} = [h_1', h_2', \dots, h_L'],
\end{equation}
where \( h_i' \in \mathbb{R}^d \) is the output representation of the \( i \)th token after being processed by routed expert.

To produce a unified representation for the entire sequence, we perform a weighted aggregation of the token outputs. The process is divided into the following steps:

\subsubsection{Weight Calculation}
For each token output \( h_i' \), a weight \( \alpha_i \) is computed to determine its contribution to the final aggregated representation:
\begin{equation}
\label{alpha_value}
\alpha_i = \frac{\exp\left(w^T h_i'\right)}{\sum_{l=1}^L \exp\left(w^T h_{l}'\right)},
\end{equation}
where \( w \in \mathbb{R}^d \) is a learnable parameter vector, and \( \alpha_i \in [0, 1] \) ensures \( \sum_{i=1}^L \alpha_i = 1 \).

\subsubsection{Weighted Aggregation}
Using the computed weights \( \alpha_i \), the token outputs are aggregated into a single representation:
\begin{equation}
h^{\text{agg}} = \sum_{i=1}^L \alpha_i h_i'.
\end{equation}

\subsubsection{Normalization and Final Transformation}
The aggregated representation \( h^{\text{agg}} \) is optionally normalized using LayerNorm, producing \( \tilde{h}^{\text{agg}} \). The normalized representation is then passed through a fully connected layer with a softmax activation function, which outputs a probability distribution over the possible outcomes for the specific task.

\subsection{Loss Functions}
To optimize the privacy-aware MoE model, we define two key loss functions: the task loss and the group-wise load balancing loss. These losses ensure that the model achieves high performance on the primary task while maintaining balanced utilization of experts within the privacy and non-privacy groups.

\subsubsection{Task Loss}
The task loss, denoted as \(\mathcal{L}_{\text{task}}\), measures the model's performance on the primary task. This loss is task-specific and can be adapted to different objectives, such as classification or sequence generation.

\subsubsection{Group-Wise Load Balancing Loss}
To ensure balanced expert utilization while preserving the privacy/non-privacy group separation, we introduce a group-wise load balancing loss. This loss operates separately within each expert group, enforcing uniform token distribution among privacy experts for sensitive tokens and among non-privacy experts for non-sensitive tokens, thereby maintaining both group isolation and computational efficiency.

To account for the separation of sensitive and non-sensitive tokens, we compute the expert usage separately for sensitive and non-sensitive tokens. Let \(L_\text{p}\) and \(L_\text{np}\) denote the number of sensitive and non-sensitive tokens, respectively. The average usage of \(K_\text{p}\) privacy experts and non-privacy experts \(K_\text{np}\) is given by:
\begin{equation}
\mathbf{u}_\text{p} = \frac{1}{L_\text{p}} \sum_{i=1}^{L} \mathbf{z}_i[:K_\text{p}],\quad \mathbf{u}_\text{np} = \frac{1}{L_\text{np}} \sum_{i=1}^{L} \mathbf{z}_i[K_\text{p}:],
\end{equation}
where \(\mathbf{u}_\text{p} \in \mathbb{R}^{K_\text{p}}\) represents the usage of privacy experts, and \(\mathbf{u}_\text{np} \in \mathbb{R}^{K_\text{np}}\) represents the usage of non-privacy experts.

The group-wise load balancing loss encourages uniform usage of experts within each group:
\begin{equation}
\mathcal{L}_\text{p} = \sum_{j=1}^{K_\text{p}} \left( u_{\text{p},j} - \frac{1}{K_\text{p}} \right)^2,
\mathcal{L}_\text{np} = \sum_{j=1}^{K_\text{np}} \left( u_{\text{np},j} - \frac{1}{K_\text{np}} \right)^2.
\end{equation}

The total group-wise load balancing loss $\mathcal{L}_{\text{lb}}$ is then given by:
\begin{equation}
\mathcal{L}_{\text{lb}} = \mathcal{L}_\text{p} + \mathcal{L}_\text{np}.
\end{equation}

\subsubsection{Total Loss}
The total loss is a weighted sum of the task loss and the group-wise load balancing loss:
\begin{equation}
\mathcal{L}_{\text{total}} = \mathcal{L}_{\text{task}} + \lambda_{\text{LB}} \mathcal{L}_{\text{lb}},
\end{equation}
where \(\lambda_{\text{LB}}\) is a hyperparameter that controls the relative importance of the load balancing loss.

\section{Bandwidth-Adaptive and Importance-Aware Token Offloading}
This section presents a bandwidth-adaptive and importance-aware token offloading scheme designed to address the challenge of transmitting non-sensitive tokens \(\mathcal{T}_\text{ns}\) under bandwidth constraints within the PWC-MoE framework. The proposed scheme incorporates an importance predictor, a neural network that estimates the importance score of each token in \(\mathcal{T}_\text{ns}\). Based on these scores, a subset of the most important tokens \(\mathcal{I}\) is selected for transmission to the remote base station, thereby adapting to the available bandwidth.

\subsection{Data Collection for Importance Prediction}
To train the importance predictor, we construct a dataset where the goal is to predict the importance scores \( \boldsymbol{\alpha} = [\alpha_1, \alpha_2, \dots, \alpha_L] \) for each token in the input sequence. These scores, as defined in (\ref{alpha_value}), represent the relative contribution of each token to the model's final output. Each token is represented by its embedding \( \mathbf{h} = [h_1, h_2, \dots, h_L] \), where \( h_i \in \mathbb{R}^d \). The ground truth importance scores \( \boldsymbol{\alpha} \) are obtained from the weighted aggregation network of the PWC-MoE model. The resulting dataset consists of input-output pairs \( \{(h_i, \alpha_i)\}_{i=1}^L \), where \( h_i \) is the token embedding and \( \alpha_i \) is the corresponding importance score.

\subsection{Architecture of the Importance Predictor}
The importance predictor takes token embeddings \( \mathbf{h} = [h_1, h_2, \dots, h_L] \) as input and predicts the corresponding importance scores \( \hat{\boldsymbol{\alpha}} \).  Its architecture includes an input layer that projects token embeddings to a lower-dimensional space to reduce computational complexity, a multi-layer transformer encoder that captures both local and long-range contextual relationships between tokens using multi-head self-attention mechanisms, and an output scoring layer with softmax normalization that produces normalized importance scores.

\subsection{Training Methodology}
The importance predictor is trained to minimize the Kullback-Leibler (KL) Divergence between the predicted importance scores \( \hat{\boldsymbol{\alpha}} \) and the ground truth \( \boldsymbol{\alpha} \). The KL divergence loss is defined as:
\begin{equation}
\mathcal{L}_{\text{KL}} = \sum_{i=1}^L \alpha_{i} \log \frac{\alpha_{i}}{\hat{\alpha}_{i}},    
\end{equation}
where \( L \) is the length of each sequence, \( \alpha_{i} \) is the ground truth importance score for the \( i \)th token, and \( \hat{\alpha}_{i} \) is the predicted importance score for the same token.

\section{Numerical Results}
In our experiments, we develop a privacy-aware MoE model by augmenting the GPT-2\cite{radford2019language} with task-specific feedforward networks. In this setup, GPT-2 is used as a pre-trained backbone and is not updated during training. The extended architecture incorporates 8 expert modules (2 privacy experts and 6 non-privacy experts), where each expert is implemented as a two-layer fully-connected network. For training and evaluation, we use the Banking77 intent classification dataset\cite{casanueva-etal-2020-efficient}, which contains 10,003 training examples and 3,080 testing examples. The model is fine-tuned on this dataset to adapt to the task-specific requirements. In our setup, any numerical tokens (e.g., account numbers or phone numbers) appearing in the dataset are treated as sensitive tokens and are handled exclusively by the privacy experts. 
The wireless simulation parameters are detailed in Table \ref{tab:wireless-parameters}.

\begin{table}[h]
\centering
\renewcommand{\arraystretch}{1.2}
\caption{Wireless system parameters}
\label{tab:wireless-parameters}
\begin{tabular}{|l|l|l|l|}
\hline
\textbf{Parameter} & \textbf{Value} & \textbf{Parameter} & \textbf{Value} \\ \hline
Carrier Freq. $f_\text{c}$ & 2.4 GHz & Bandwidth $W_\text{c}$ & 10 MHz \\ \hline
Commu. Time $T^{\text{ul}}$ & 100 ms & Client Tx Power $P$ & 23 dBm \\ \hline
PSD $N_0$ & -174 dBm/Hz & Std. Dev. $\sigma$ & 7.8 dB \\ 
\hline
\end{tabular}
\end{table}

\begin{figure}[htbp]
\centerline{\includegraphics[scale=0.47]{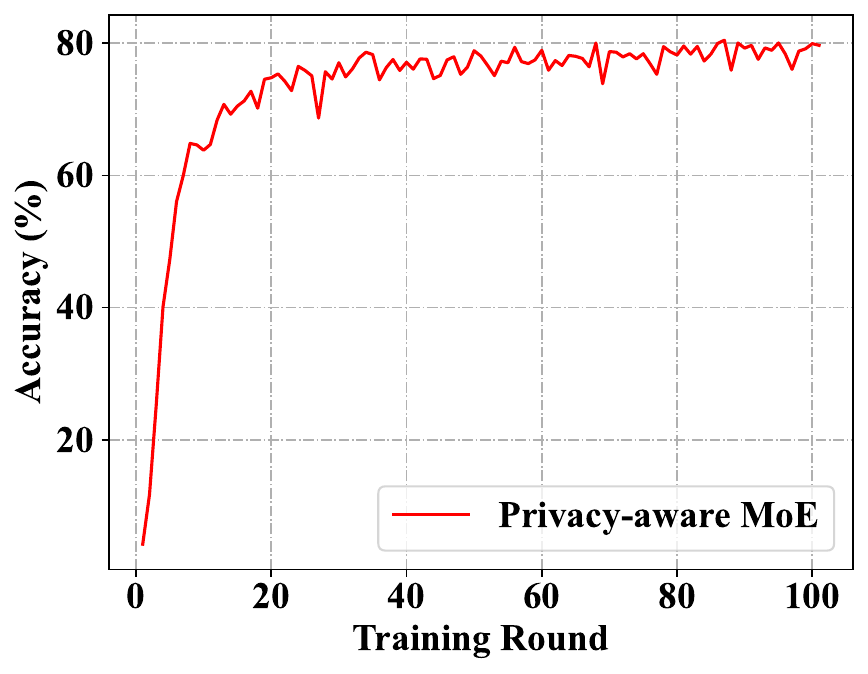}}
\caption{Accuracy with training rounds.}
\label{training_accuracy}
\end{figure}

Fig.~\ref{training_accuracy} illustrates the accuracy convergence of the privacy-aware MoE model during the training process at the server side, reflecting the model's performance when processing all tokens. The x-axis represents the training rounds, while the y-axis shows the accuracy on the test set. The model's accuracy improves steadily with the number of training rounds and converges around the 40th round, eventually stabilizing at approximately 78\% accuracy. This indicates that the model effectively adapts to the task requirements during training.

\begin{figure}[htbp]
\centerline{\includegraphics[scale=0.30]{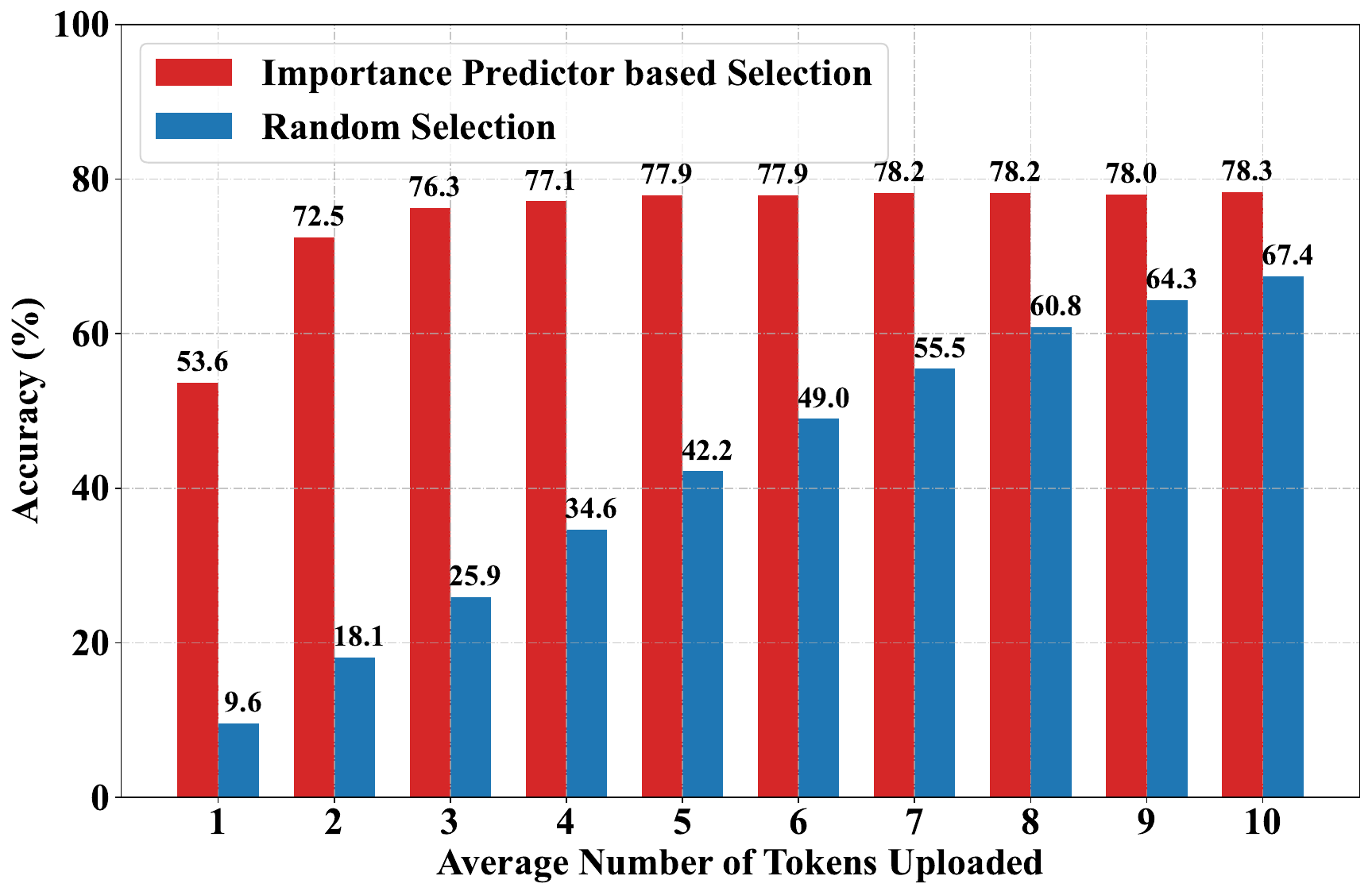}}
\caption{Accuracy comparison with different numbers of uploaded tokens.}
\label{predictor_vs_random}
\end{figure}

\begin{figure*}[htbp]
    \centering
    \subfigure[]{
        \includegraphics[scale=0.29]{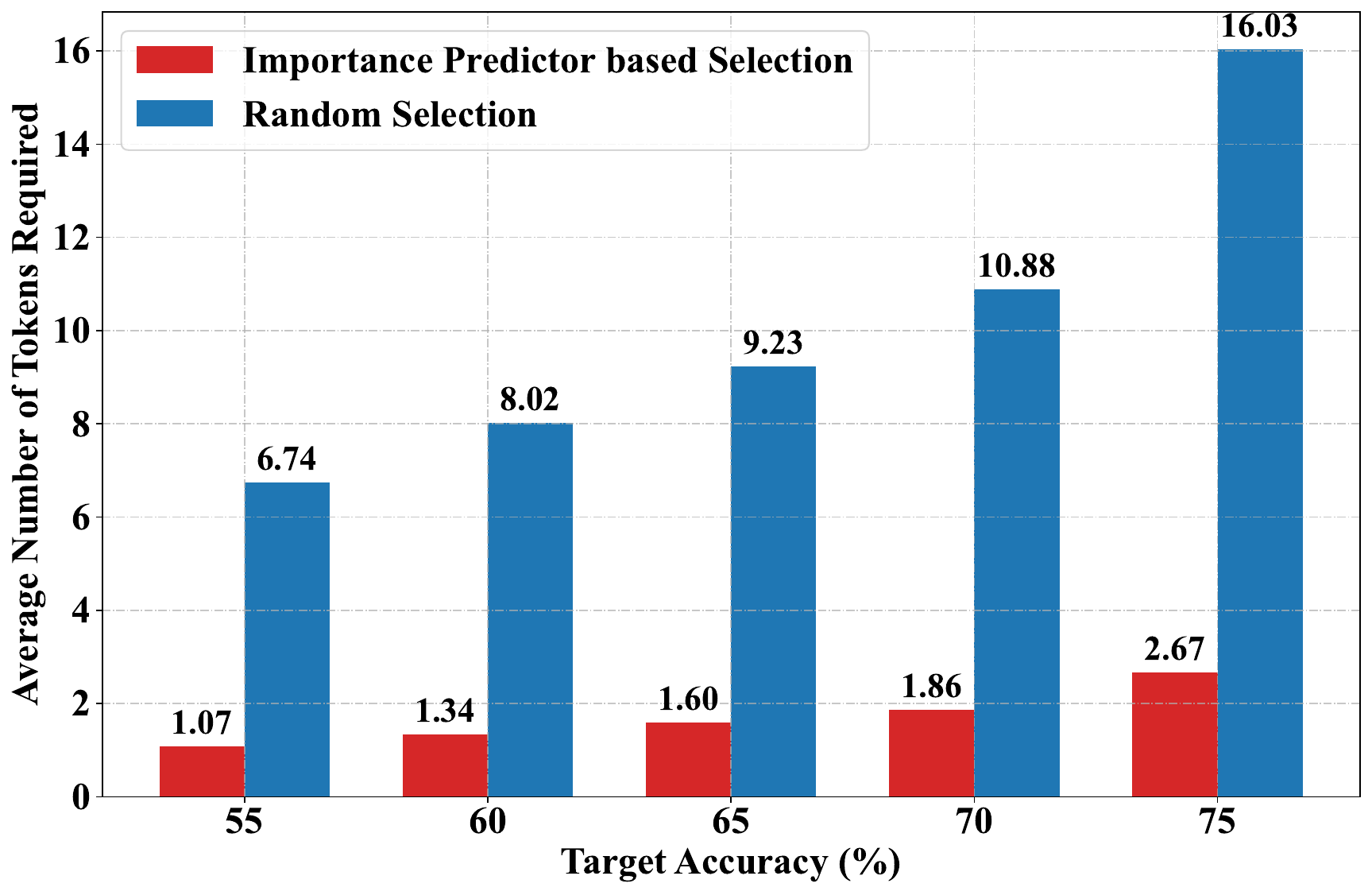}
        \label{tokens_required_by_accuracy}
    }
    \subfigure[]{
        \includegraphics[scale=0.29]{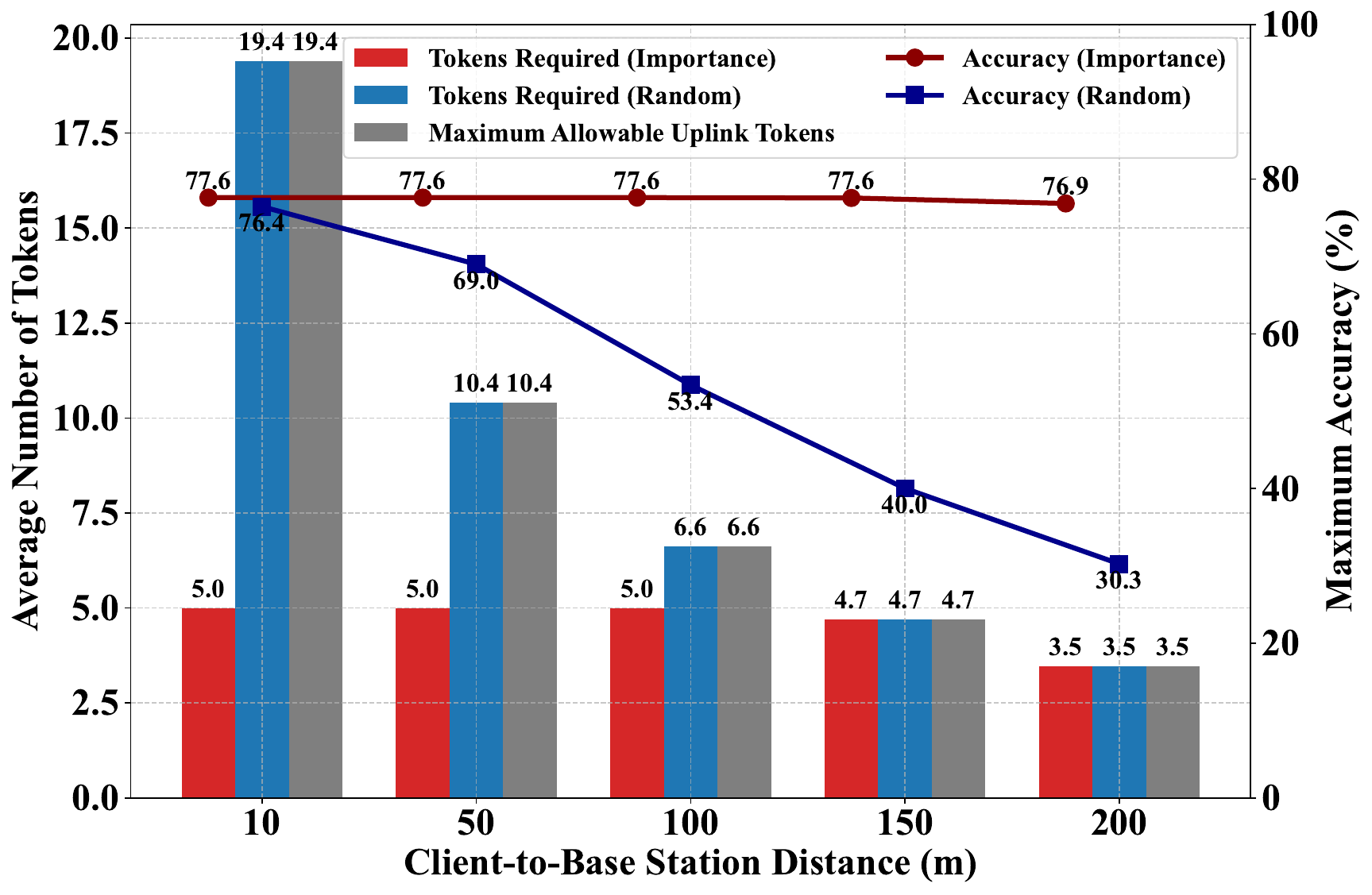}
        \label{tokens_vs_max_tokens_distance}
    }
    \caption{(a) The number of tokens required to meet the target accuracy. (b) Minimum number of tokens required to achieve maximum accuracy.}
    \label{fig:average_tokens}
\end{figure*}

Fig.~\ref{predictor_vs_random} illustrates the accuracy comparison of the PWC-MoE framework under different token selection schemes, where the x-axis represents the number of tokens transmitted per example in the test set, constrained by the wireless channel conditions. 
The red bars represent the performance of the importance predictor-based selection method, while the blue bars correspond to the random selection method. It can be observed that the importance predictor-based method consistently outperforms the random selection method for the same number of uploaded tokens. Specifically, the accuracy of the importance predictor-based method reaches 77.9\% when 5 tokens are uploaded, and beyond this point, the accuracy stabilizes around 78\% with minimal fluctuations. In contrast, the random selection method shows a linear increase in accuracy as the number of uploaded tokens increases, but even with 10 tokens, it only achieves an accuracy of 67.4\%, which is still lower than the performance of the importance predictor-based method. This demonstrates the efficiency and effectiveness of the importance predictor-based selection in utilizing limited transmission resources.

Fig.~\ref{fig:average_tokens}~\subref{tokens_required_by_accuracy}  illustrates the average number of tokens required per example in the test set to achieve the target accuracy for both the importance predictor-based selection method and the random selection method. Across all accuracy levels, the importance predictor-based method consistently requires fewer tokens to achieve the same target accuracy compared to the random selection method. 
Fig.~\ref{fig:average_tokens}~\subref{tokens_vs_max_tokens_distance} illustrates the minimum required tokens for each selection method to achieve their respective maximum accuracy under varying client-to-base station distances, where the maximum allowable uplink tokens decrease as the distance increases (gray bars). Both selection methods are constrained by this token limit. The random selection method typically requires close to the maximum allowable tokens to reach its best performance, while the importance predictor-based method achieves comparable or higher accuracy with fewer tokens.

\section{Conclusion}
In this paper, we proposed a PWC-MoE framework to address the challenges of privacy preservation, performance optimization, and bandwidth efficiency in deploying LLMs. The framework dynamically routes sensitive tokens to local privacy experts while offloading non-sensitive tokens to remote experts, ensuring both privacy and computational efficiency. We further proposed bandwidth-adaptive and importance-aware token offloading mechanism prioritizes the transmission of critical tokens based on their importance scores and available bandwidth, effectively reducing communication costs. Experiments demonstrated that the framework achieves high accuracy, efficient resource utilization, and adaptability to varying bandwidth conditions, making it a practical solution for deploying LLMs in privacy-sensitive and resource-constrained environments.

%%%%%%%%%%%%%%%%%%%%%%%%%%%%%%%%%%%%%%%%%%%%%%%%%%%%%%%%%%%%%%%%%%%%%%%%

%%% The next two lines define, first, the bibliography style to be 
%%% applied, and, second, the bibliography file to be used.
\bibliographystyle{IEEEtran}
\bibliography{IEEEabrv,mylib}

\begin{thebibliography}{10}
\providecommand{\url}[1]{#1}
\csname url@samestyle\endcsname
\providecommand{\newblock}{\relax}
\providecommand{\bibinfo}[2]{#2}
\providecommand{\BIBentrySTDinterwordspacing}{\spaceskip=0pt\relax}
\providecommand{\BIBentryALTinterwordstretchfactor}{4}
\providecommand{\BIBentryALTinterwordspacing}{\spaceskip=\fontdimen2\font plus
\BIBentryALTinterwordstretchfactor\fontdimen3\font minus \fontdimen4\font\relax}
\providecommand{\BIBforeignlanguage}[2]{{%
\expandafter\ifx\csname l@#1\endcsname\relax
\typeout{** WARNING: IEEEtran.bst: No hyphenation pattern has been}%
\typeout{** loaded for the language `#1'. Using the pattern for}%
\typeout{** the default language instead.}%
\else
\language=\csname l@#1\endcsname
\fi
#2}}
\providecommand{\BIBdecl}{\relax}
\BIBdecl

\bibitem{zhao2023survey}
W.~X. Zhao, K.~Zhou, J.~Li, T.~Tang, X.~Wang, Y.~Hou, Y.~Min, B.~Zhang, J.~Zhang, Z.~Dong \emph{et~al.}, ``{A survey of large language models},'' \emph{arXiv preprint arXiv:2303.18223}, Mar. 2023.

\bibitem{van2024survey}
C.~Van~Nguyen, X.~Shen, R.~Aponte, Y.~Xia, S.~Basu, Z.~Hu, J.~Chen, M.~Parmar, S.~Kunapuli, J.~Barrow \emph{et~al.}, ``A survey of small language models,'' \emph{arXiv preprint arXiv:2410.20011}, Oct. 2024.

\bibitem{jiao-etal-2020-tinybert}
X.~Jiao, Y.~Yin, L.~Shang, X.~Jiang, X.~Chen, L.~Li, F.~Wang, and Q.~Liu, ``{T}iny{BERT}: Distilling {BERT} for natural language understanding,'' in \emph{Find. Assoc. Comput. Linguist.: EMNLP 2020}, Nov. 2020, pp. 4163--4174.

\bibitem{zhang2024llm}
K.~Zhang, L.~Peng, C.~Wang, A.~Go, and X.~Liu, ``Llm cascade with multi-objective optimal consideration,'' \emph{arXiv preprint arXiv:2410.08014}, Oct. 2024.

\bibitem{cai2025survey}
W.~Cai, J.~Jiang, F.~Wang, J.~Tang, S.~Kim, and J.~Huang, ``A survey on mixture of experts in large language models,'' \emph{IEEE Trans. Knowl. Data Eng.}, Mar. 2025.

\bibitem{3gpp_tr_38_901_v16_1_0}
{3rd Generation Partnership Project (3GPP)}, ``Technical specification group radio access network; study on channel model for frequencies from 0.5 to 100 ghz (release 16),'' \emph{3GPP TR 38.901 V16.1.0}, Dec. 2019.

\bibitem{akbik2019pooled}
A.~Akbik, T.~Bergmann, and R.~Vollgraf, ``Pooled contextualized embeddings for named entity recognition,'' in \emph{Proc. Conf. North Am. Chapter Assoc. Comput. Linguist. (NAACL)}, Jun. 2019, pp. 724--728.

\bibitem{DBLP:conf/iclr/JangGP17}
E.~Jang, S.~Gu, and B.~Poole, ``{ Categorical Reparameterization with Gumbel-Softmax },'' in \emph{Proc. Intern. Conf. Learn. Represent. (ICLR)}, Apr. 2017.

\bibitem{radford2019language}
A.~Radford, J.~Wu, R.~Child, D.~Luan, D.~Amodei, I.~Sutskever \emph{et~al.}, ``Language models are unsupervised multitask learners,'' \emph{OpenAI blog}, vol.~1, no.~8, p.~9, Feb. 2019.

\bibitem{casanueva-etal-2020-efficient}
I.~Casanueva, T.~Tem{\v{c}}inas, D.~Gerz, M.~Henderson, and I.~Vuli{\'c}, ``Efficient intent detection with dual sentence encoders,'' in \emph{Proc. 2nd Worksh. Nat. Lang. Process. Convers. AI (NLP4ConvAI)}, Jul. 2020, pp. 38--45.

\end{thebibliography}

%%%%%%%%%%%%%%%%%%%%%%%%%%%%%%%%%%%%%%%%%%%%%%%%%%%%%%%%%%%%%%%%%%%%%%%%

\end{document}